\documentclass[journal]{IEEEtran}
\IEEEoverridecommandlockouts
\usepackage{cite}
\usepackage{amsmath,amssymb,amsfonts}
\usepackage{algorithmic}
\usepackage{lipsum,adjustbox}
\usepackage{verbatim}
\usepackage{graphics}
\usepackage{nicefrac}
\usepackage{stackengine}
\usepackage{pgfplots}
\usepackage{tikz}
\pgfplotsset{width=7cm,compat=1.8}
\usepackage{multirow}
\usepackage{soul}
\usepackage{amssymb}
\usepackage{amsmath}
\usepackage{algorithm}
\usepackage{algorithmic}
\usepackage{graphicx}
\usepackage[caption=false]{subfig}
\usepackage{textcomp}
\usepackage{booktabs}

\usepackage{stfloats}
\usepackage{graphicx}

\usepackage{scalefnt}
\usepackage{mathrsfs}         
\def\BibTeX{{\rm B\kern-.05em{\sc i\kern-.025em b}\kern-.08em
    T\kern-.1667em\lower.7ex\hbox{E}\kern-.125emX}}
\usepackage[nolist]{acronym}
\begin{acronym}

    \acro{UAV}{unmanned aerial vehicle}
    \acro{MARL}{multi-agent reinforcement learning}
    \acro{IoT}{internet of things}
    \acro{LPWAN}{low-power wide area networks}
    \acro{DQN}{Deep Q Network}
    \acro{A2C}{Advantage Actor Critic}
    \acro{POMDP}{Partially Observable Markov Decision Process}
    \acro{MADDPG}{multi-agent deep deterministic policy gradient}
    \acro{MAVEN}{multi-agent variational exploration}
    \acro{MF-QL}{mean field Q learning}
    \acro{MF-AC}{mean field actor critic}
    \acro{COMA}{counterfactual multi-agent policy Gradients}
    \acro{DGN}{graph convolutional reinforcement learning}
    \acro{MAGNet}{}      
    \acro{HAMA}{}
    \acro{G2ANet}{}
    \acro{RIAL}{reinforced inter-agent learning}
    \acro{DIAL}{differentiable inter-agent learning}
    \acro{CommNet}{}
    \acro{BiC-Net}{}
    \acro{A3C2}{asynchronous advantage actor critic with communication}
    \acro{ATOC}{attentional communication model}
    \acro{TarMAC}{targeted multi-agent communication}
    \acro{I2C}{individually inferred communication}
    \acro{IC3Net}{individualized controlled continuous communication model}
    \acro{SchedNet}{}
    \acro{IS}{intention sharing}
    \acro{VBC}{variance based control}
    \acro{AI}{artificial intelligence} 
    \acro{EC}{emergent communication}
    \acro{MAAC}{multi actor attention critic}
    \acro{TarMAC}{targeted multi-agent Communication for collaborative multi-agent deep reinforcement learning}
    \acro{I2C}{individually inferred communication}
    \acro{A3C2}{asynchronous advantage actor critic with communication}  
    \acro{CTDE}{centralized learning and decentralized execution}
    \acro{RL}{reinforcement learning}
\end{acronym}

\usepackage[utf8]{inputenc}
\usepackage[english]{babel}
\usepackage{amsthm}
\usepackage{commath}
\usepackage{multicol}
\usepackage{pdflscape}
\usepackage{comment}
\usepackage{siunitx}
\usepackage{hyperref}

\usetikzlibrary {shadows}
\usetikzlibrary{shadows.blur}
\usetikzlibrary{shadows.blur,positioning}
\begin{document}

\title{
Emergent Communication in Multi-Agent Reinforcement Learning for Future Wireless Networks
\thanks{

M. Chafii is with the Engineering Division, New York University (NYU) Abu Dhabi, 129188, UAE and NYU WIRELESS, NYU Tandon School of Engineering, Brooklyn, 11201, NY (e-mail: marwa.chafii@nyu.edu). S. Naoumi is with NYU Tandon School of Engineering, NY.
R. Alami, E. Almazrouei and M. Debbah are with Technology Innovation Institute (TII), Abu Dhabi, UAE.
M. Bennis is with the Centre for Wireless Communications, the University
of Oulu, Finland. This project received funding from TII Abu Dhabi.
}
}

 \author{Marwa Chafii,
          Salmane Naoumi,
         Reda Alami,
                  Ebtesam Almazrouei,
                  Mehdi Bennis,~\IEEEmembership{Fellow,~IEEE}
         and~Merouane Debbah,~\IEEEmembership{Fellow,~IEEE}}

\maketitle

\begin{abstract}
In different wireless network scenarios, multiple network entities need to cooperate in order to achieve a common task with minimum delay and energy consumption. Future wireless networks mandate exchanging high dimensional data in dynamic and uncertain environments, therefore implementing communication control tasks becomes challenging and highly complex. Multi-agent reinforcement learning with emergent communication (EC-MARL) is a promising solution to address high dimensional continuous control problems with partially observable states in a cooperative fashion where agents build an emergent communication protocol to solve complex tasks. 
This paper articulates the importance of EC-MARL within the context of future 6G wireless networks, which imbues autonomous decision-making capabilities into network entities to solve complex tasks such as autonomous driving, robot navigation, flying base stations network planning, and smart city applications. An overview of EC-MARL algorithms and their design criteria are provided while presenting use cases and research opportunities on this emerging topic.
\end{abstract}


\section{Introduction} \label{introduction}
\IEEEPARstart{A} {s}
 we are in the early phase of the development of the 6th generation (6G) of wireless networks, researchers from academia and industry are investigating what 6G will be \cite{10041914}. There is a common consensus that at least two dimensions will characterize 6G, namely sensing and \ac{AI}.
With AI-enabled devices that perform both sensing and communication, processing and communicating a large amount of data becomes challenging. The post-Shannon communication paradigm allows for efficient encoding and transmission of semantic information. Instead of guaranteeing the correct reception of data symbols regardless of their meaning and semantics, the post-Shannon paradigm focuses on the meaning of data and how they affect specific actions for downstream tasks.
One approach for addressing such a problem is to transmit only relevant data samples that are decoded at the receiver. The relevance of data refers to their meaning and/or their importance in achieving a certain goal. Another approach concerns communication between intelligent agents, where communication is no longer hard coded into a classical vocabulary but the message representation is learned through interaction. In this case, semantics and their task-oriented encoding emerge from play and interaction. This paradigm is referred to as emergent communication.

The problem of intelligent agents interacting with an environment to collaboratively complete a task can be solved within the framework of \ac{MARL}, where \ac{RL} is used by each agent to interact with the environment and learn a policy to maximize its cumulative reward. A large number of telecommunications scenarios in sensor networks, \ac{IoT} networks, traffic control, resource management, and smart factories applications can be represented within this framework.

Oftentimes, interaction with an environment may not be enough, in which agents need to communicate with each other to solve a common task. In fact, communications in MARL are needed in the following cases i) partially observable environment: each agent has only access to its observation and not to the current state of the whole environment, so in case of common reward among agents, they need to exchange information about their own observation which is not always available for other agents ii) non-stationarity of the environment: from each agent's perspective, since the resulting reward from an action taken by an agent depends on the decision making of other agents, in case of common reward, it is in the best interest of the agents to share their decision making policies with other agents. Moreover, the physical environment can also be rapidly changing, hence agents need to communicate to compensate for their partial knowledge of the environment by exchanging their observations, policies and experience, which allows the agents to coordinate toward the execution of the common task. Without communication, agents and their decisions are considered part of the environment from an agent's perspective, and classical MARL algorithms show poor performance and usually fail to converge to an optimal policy.

In this paper, we motivate the importance of the \ac{EC}-\ac{MARL} framework in solving 6G use cases in an environment characterized by intelligent agents, high-dimensional sensing data and limited communications resources.
We show why emergent communication is more efficient than being hard-coded in future wireless networks, and present the general framework of \ac{EC}-\ac{MARL}, with the most popular algorithms and their design criteria. Moreover, we define use cases and present challenges and open directions for the research community.

\section{Why EC-MARL for 6G Communications?}
It is expected that 6G networks will feature communication devices with AI and sensing abilities. When a large number of such devices are deployed, communication between them can be limited because of the following constraints: i) energy constraints: when communicating agents refer to energy-constrained devices such as \ac{LPWAN} devices, sensors and \acp{UAV} that run on a limited lifetime battery, the amount of data agents can exchange with each other is limited; ii) low latency requirements: in some applications such as autonomous driving and \ac{UAV} swarm control, exchanging information in a timely manner between agents is important. Encoding large observations in short messages would contribute to reducing communication latency;
iii) spectrum limitation: When a large number of sensors is deployed or high dimensional sensing information is involved e.g. 3D videos, High definition (HD) images, exchanging raw sensing data requires large bandwidths and may not be affordable when a large number of agents is transmitting.

To tackle these communication constraints in multi-agent systems, agents need to exchange only the \emph{relevant} information to solve the task instead of transmitting the whole observation as is the case in classical MARL systems. 
However, in complex environments where different parameters are involved, it is challenging to analytically extract the relevant information, as the agent is not aware of which part of the sensing data is correlated with the task to be completed, and which part of the data would be helpful for other agents, due to the partially observable environment from each agent's perspective. However, this relevant information can be learned by interaction with other agents i.e., instead of transmitting fixed communication messages, their goal-oriented representations that encode the most relevant useful information from the raw sensing data can be learned. 

Unlike communication protocols that are manually coded and exchanged between multiple agents to coordinate for executing some task, EC-MARL considers algorithms that learn both policies and communication protocols by each agent. Each agent takes an action and learns a communication message based on his observation as well as communication messages transmitted by other agents. Transmitted messages are not hard coded or specified but learned. 
This learned communication has been referred to as the emergence of communication \cite{lazaridou2020emergent}.

Emergence of communication in MARL systems can be studied within the new paradigm of semantic and goal-oriented communications \cite{seo2021semantics}. At the beginning of the simulation, agents exchange dummy messages without any meaning. Through interaction, semantics and meanings emerge from communication messages \cite{lazaridou2020emergent}, i.e. agents do not transmit pre-determined messages, but learn a task-oriented message by encoding the most meaningful information to the execution of the task at hand. Nevertheless, it is not always easy or possible to understand the meaning of the learned message by an external observer, decoding emergent communication protocols is currently an active research area.

\section{EC-MARL Framework}
EC-MARL is an application of \ac{MARL} in partially-observable Markov games, in environments where agents have a joint communication channel. In every state, agents take actions given partial observations of the true world state, including messages sent on a shared channel, and each agent obtains an individual reward after interacting with the environment. Through their individual experiences interacting with one another and with the environment, agents learn to broadcast appropriate messages, interpret messages received from peers, and act accordingly.
Formally, a multi-agent system considers an $N$-player partially observable Markov game $G$ defined on a finite state set $\mathcal{S}$, with action sets $\left(\mathcal{A}^{1}, \ldots, \mathcal{A}^{N}\right)$ and message sets $\left(\mathcal{M}^{1}, \ldots, \mathcal{M}^{N}\right)$. An observation function $O: \mathcal{S} \times\{1, \ldots, N\} \rightarrow \mathbb{R}^{d}$ defines each agent's $d$-dimensional restricted view of the true state space. On each time-step $t$, each agent $i$ receives as an observation $o_{t}^{i}=O\left(\mathcal{S}_{t}, i\right)$, and the messages $m_{t-1}^{j}$ sent in the previous state for all $j \neq i$. Each agent $i$ then selects an environment action $a_{t}^{i} \in \mathcal{A}^{i}$ and a message action $m_{t}^{i} \in \mathcal{M}^{i}$. Every agent gets an individual reward $r_{t}^{i}: \mathcal{S} \times \mathcal{A}^{1} \times \cdots \times \mathcal{A}^{N} \rightarrow \mathbb{R}$ for player $i$. 
In the fully cooperative setting, each agent receives the same reward at each timestep, $r_{t}^{i}=$ $r_{t}^{j} \quad \forall i, j \leq N$, which is denoted by $r_{t}$. Each agent maintains an action and a message policy from which actions and messages are sampled, $a_{t}^{i} \sim \pi_{A}^{i}\left(\cdot \mid x_{t}^{i}\right)$ and $m_{t}^{i} \sim \pi_{M}^{i}\left(\cdot \mid x_{t}^{i}\right)$, and which can be in general functions of their entire trajectory of experience $x_{t}^{i}:=\left(\mathbf{m}_{0}, o_{1}^{i}, a_{1}^{i}, \ldots, a_{t-1}^{i}, \mathbf{m}_{t-1}, o_{t}^{i}\right)$, where $\mathbf{m}_{\mathbf{t}}=\left(m_{t}^{1}, \ldots, m_{t}^{N}\right)$. These policies are optimized to maximize discounted cumulative joint reward $J\left(\boldsymbol{\pi}_{A}, \boldsymbol{\pi}_{M}\right):=$ $\mathbb{E}_{\boldsymbol{\pi}_{A}, \boldsymbol{\pi}_{M}, \mathcal{T}}\left[\sum_{t=1}^{\infty} \gamma^{t-1} r_{t}\right]$ (which is discounted by $\gamma<1$ to ensure convergence), where $\pi_{A}:=$ $\left\{\pi_{A}^{1}, \ldots, \pi_{A}^{N}\right\}, \pi_{M}:=\left\{\pi_{M}^{1}, \ldots, \pi_{M}^{N}\right\}$.

\begin{figure*}
  \input{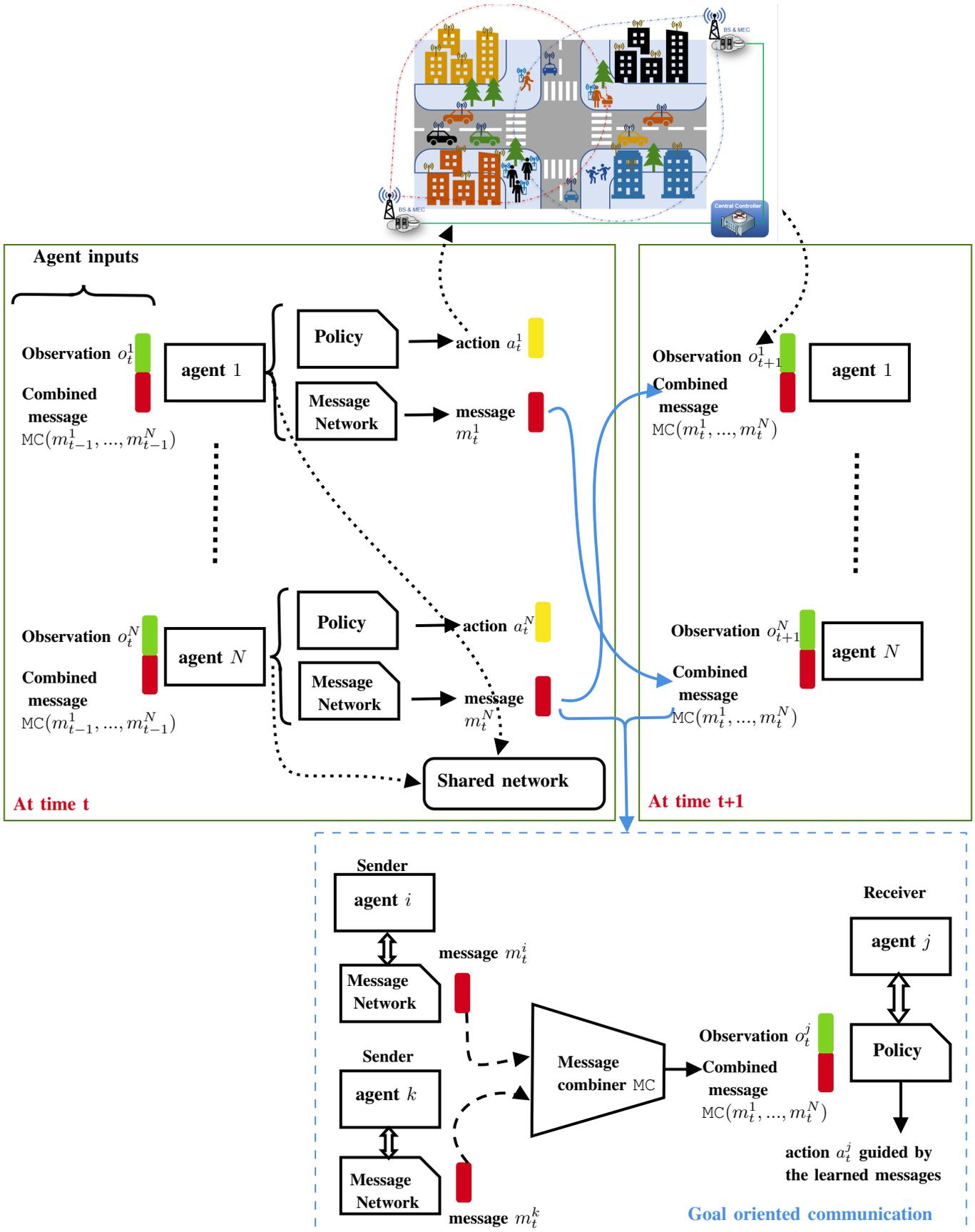}
  \caption{Emergent communication in multi-agent reinforcement learning (EC-MARL). Agents learn new communication protocols and how to transmit relevant messages to other agents in order to solve complex tasks. 
  }
  \label{fig.Framework}
\end{figure*}

The objective $J\left(\pi_{A}, \pi_{M}\right)$ is a joint objective, thus the learning model is that of a centralized learning and decentralized execution, where every agent has its own experience in the environment and using the experiences of the other agent, it optimizes the objective $J$ with respect to its own action and message policies $\pi_{A}^{i}$ and $\pi_{M}^{i}$; there is no direct communication between agents other than using the actions and message channel in the environment.
It should be noted that applying independent reinforcement learning to cooperative Markov games results in a problem for each agent which is non-stationary and non-Markov, and presents difficult joint exploration and coordination problems. Thus, emergent communication becomes crucial to handle the perfect coordination between the agents. Figure~ \ref{fig.Framework} summarizes the framework of learning communication protocol for an EC-MARL algorithm. 

\section{EC-MARL Algorithms} \label{sec:Formalism}

\subsection{Designing EC-MARL Algorithms}
In the literature \cite{zhu2022survey}, we find different EC-MARL algorithms that are designed differently mainly according to the following criteria:

\paragraph{To whom?} each agent needs to define to whom he needs to transmit the message. The agent can
\begin{itemize}
    \item Broadcast the message to all agents;
    \item Multicast the message to specific agents for example its closest neighbors in terms of distance, quality of the channel, or agents to whom the information is relevant. Agents can also specify to target a specific group of agents e.g. agents located on the first floor;
    \item Unicast the message to a specific agent.
\end{itemize}
In some scenarios, agents would communicate through a relay agent that would gather the messages and relay them to target agents. This case can also be considered as a unicast transmission. Note that the network of agents can be heterogeneous where each agent has a different communication type, which can also change over time depending on the message to communicate and the context of communication.
\paragraph{What?} agents need to decide what information to exchange
\begin{itemize}
    \item Existing context: agents encode their past observations and actions into messages;
    \item Predicted context: agents encode their intended action and predicted future actions e.g. car intends to brake.
\end{itemize}
A combination of both can also be encoded and communicated to other agents.
\paragraph{How?} agent receiving messages from other agents can
\begin{itemize}
    \item Concatenate the messages;
    \item Apply linear combination: e.g. summation, averaging. A weighted linear combination of messages can give more value to messages of specific agents. For example in search and rescue missions, messages from neighboring agents are more relevant.
\end{itemize} 
\paragraph{Where?} 
where is the communication integrated in the learning process? The message can be seen as an additional observation that the agent can take as an extra input to the value function, or to the policy function or to both. 
\begin{itemize}
    \item Value level: most works based on Deep-Q-network algorithm take the messages as input into the value function or the critic network\cite{foerster2016learning, IntensionSharing}. 
    \item Policy level: messages can also be incorporated into the policy function or the actor-network by conditioning the agents' next action on the received messages. Each agent will no longer act independently by exploiting information from other agents. 
    \item Policy level and value level: another way to incorporate the messages is to consider them as extra input for both the policy model and the value model. The messages can also be combined with the observations to generate new internal states to be considered for both the actor and the critic networks.
\end{itemize}

\paragraph{When?} each agent needs to decide whether to transmit a message or skip communications. The decision may be based on the observation and whether there is relevant information to transmit. This decision can be learned as well. 
Agents can also decide to communicate in multiple rounds before taking action.

\paragraph{Training scheme}
\begin{itemize}
    \item Centralized learning: the central unit receives experiences from all agents and learns actions. Most EC-MARL works do not consider this scheme;
    \item Decentralized learning: agents have independent training processes;
    \item \Ac{CTDE}: communication between agents is not restricted during learning, which is performed by a centralized algorithm; however, during the execution of the learned policies, agents can communicate only via a limited-bandwidth channel. Indeed,  each agent $i$ holds an individual policy that maps local observations to a distribution over individual actions. During training, agents are endowed with additional information, which is then discarded at test time.
\end{itemize}

Note that algorithms should also take into consideration communication constraints: i) Limited resources: mainly limited spectrum and energy: algorithms can consider short messages e.g. RIAL and DIAL transmit 1-symbol messages \cite{foerster2016learning}. We can also allow only a subset of agents to transmit their messages depending on their relevance because of the limited time and frequency and spatial resources. ii) Noisy channel: algorithms should be designed to take into consideration the noise introduced by the communication medium. Moreover, communication messages can take the form of continuous vectors, or discrete ones using a sequence of symbols.

In Table~\ref{tb:SoA_Summary_Emergent}, we summarize some popular EC-MARL algorithms with respect to the presented design criteria. All the presented algorithms follow the CTDE training scheme. Note that, the presented algorithms are tested in different simulation environments, and a fair comparison in terms of performance and complexity needs to be conducted by testing all algorithms in the same simulation environment.

\subsection{Overview of EC-MARL Algorithms}

In the literature of EC-MARL algorithms, we find the \ac{DIAL} \cite{foerster2016learning} which is based on the combination of centralized learning and Q-networks that make it possible, not only to share parameters but to push gradients from one agent to another through the communication channel. Thus, while the \ac{RIAL} is end-to-end trainable within each agent, DIAL is end-to-end trainable across agents. Letting gradients flow from one agent to another gives them richer feedback, reducing the required amount of learning by trial and error, and easing the discovery of effective protocols.

CommNet algorithm \cite{zhu2022survey} 
is based on a continuous communication channel i.e. continuous vector, which allows training using back-propagation. It uses a single network for all agents where multiple communication messages are broadcast between agents at each time step to decide the actions of all agents. The algorithm does not allow distributed execution, and assumes perfect communication conditions. 

\Ac{MAAC} \cite{actorAttentionCritic}) is an attention-based actor-critic algorithm where an attention model is learned to share the information between the policies. Moreover, this approach is able to train policies in environments with any reward setup and different action spaces for each agent.

\Ac{TarMAC} \cite{Tarmac} allows each individual agent to actively select which other agents to address messages to. This targeted communication behavior is operationalized via a simple signature-based soft attention mechanism: along with the message, the sender broadcasts a key that encodes properties of agents the
message is intended for, and is used by receivers to gauge
the relevance of the message. 

\Ac{I2C} \cite{zhu2022survey}
is a simple effective model to allow agents to learn a prior for agent-agent communication. Indeed, each agent exploits its learned prior knowledge to understand which agent is relevant and influential by just local observation. The learning of the prior knowledge is done through causal inference via a feed-forward neural network that maps its local observation to a belief about who to communicate with.

In \cite{SchedNet}, a novel architecture that incorporates an intelligent scheduling entity in order to facilitate inter-agent communication in both limited-bandwidth and shared medium access scenarios has been proposed. The architecture is called "SchedNet" which consists of an actor-network, a scheduler and a critic network.

\Ac{A3C2} \cite{A3C2} is an architecture allowing to learn messages between agents via a message neural network. A3C2 can be seen as an augmentation of the classical A3C by adding a neural network message to each agent.  The objective function of the message neural network is directly linked to the policy network.
by the message neural network.

MAGIC \cite{zhu2022survey} 
is an EC-MARL architecture based on a
graph-attention communication protocol in which a scheduler is used to communicate and whom to address messages to, using a message processor through
graph attention networks (GATs) with dynamic graphs to handle communication signals. The scheduler consists of a graph
attention encoder and a differentiable attention mechanism, which
outputs dynamic, differentiable graphs to the message processor,
which enables the scheduler and message processor to be trained
end-to-end.

For more details on EC-MARL algorithms, the interested reader can refer to the survey in \cite{zhu2022survey}.

\subsection{Lessons learned from EC-MARL Literature Review}

The literature review on \ac{EC}-\ac{MARL} algorithms reveals a diverse and innovative set of approaches to facilitate communication among agents. Key lessons include the importance of allowing gradients to flow between agents, as seen in DIAL, to reduce learning through trial and error and enhance the discovery of effective communication protocols. Algorithms like CommNet utilize continuous communication channels for synchronous execution, while others, like TarMAC, enable targeted communication via attention mechanisms. The incorporation of scheduling entities, such as in SchedNet, provides a solution for communication in limited-bandwidth scenarios. Approaches like I2C demonstrate the power of local observation for determining communication needs. MAGIC, utilizing graph attention networks, demonstrates a flexible architecture that adapts to communication demands dynamically. Across these methodologies, there is an underlying theme of striving for balance between centralized control and distributed execution, the implementation of attention mechanisms, and the use of graph-based models to build more robust and effective communication protocols. This body of work highlights the breadth of strategies available to enhance collaboration and adaptability in multi-agent systems, emphasizing the centrality of efficient and effective communication.

\subsection{Distributed Coordination in EC-MARL}

All reviewed algorithms are based on centralized execution, potentially limiting scalability and robustness in certain applications.
In a multi-agent network, distributed coordination is critical to enable agents to act independently without relying on a central entity. Distributed algorithms can enhance efficiency, fault tolerance, and adaptability to varying environments.
To achieve distributed algorithms in EC-MARL, several approaches can be considered such as local communication protocols, decentralized training, swarm intelligence techniques, or hybrid centralized-decentralized techniques.

\begin{center}
    
\begin{table*}[t]
\caption{Main deep reinforcement learning algorithms with emergent communication.}
\label{tb:SoA_Summary_Emergent}
\centering
\begin{tabular}{|c|c|c|c|c|c|c|c|c|c|c|}
\hline
Architecture &    
To whom  & 
What 
&
How  & 

\begin{tabular}[c]{@{}c@{}}Where \end{tabular}   &
When & 
\begin{tabular}[c]{@{}c@{}}Tested \\ scenarios \end{tabular}
  &  \begin{tabular}[c]{@{}c@{}}Available \\ code\end{tabular} & Limitations
  \\ \hline

RIAL \cite{foerster2016learning}
&  Broadcast  & \begin{tabular}[c]{@{}c@{}}Existing\\ context \end{tabular}  & Concatenation  & Value level &
\begin{tabular}[c]{@{}c@{}}One \\ round \end{tabular}  & 
\begin{tabular}[c]{@{}c@{}} Switch riddle \\ MNIST games  \end{tabular}    & - & \begin{tabular}[c]{@{}c@{}} no gradients are \\ passed between agents \end{tabular}  \\ \hline

DIAL \cite{foerster2016learning}
&    Broadcast  & \begin{tabular}[c]{@{}c@{}}Existing\\ context \end{tabular}  & Concatenation  & Value level   &  
 \begin{tabular}[c]{@{}c@{}}One \\ round \end{tabular} & 
\begin{tabular}[c]{@{}c@{}} Switch riddle \\ MNIST games  \end{tabular}  & - & Discrete messages\\ \hline

CommNet \cite{zhu2022survey} 
&   Broadcast  & \begin{tabular}[c]{@{}c@{}}Existing\\ context \end{tabular}  & \begin{tabular}[c]{@{}c@{}}Linear \\ combination \end{tabular}  & Policy level
& \begin{tabular}[c]{@{}c@{}}One \\ round \end{tabular}
&  \begin{tabular}[c]{@{}c@{}} Combat task \\ bAbI toy QA \\Traffic-junction \end{tabular}   & \href{https://github.com/facebookarchive/CommNet}{CommNet Code} & Scalability \\ \hline

TarMAC \cite{Tarmac}
&  Broadcast  & \begin{tabular}[c]{@{}c@{}}Existing\\  context \end{tabular}  & \begin{tabular}[c]{@{}c@{}}Weighted\\ linear \\ combination \end{tabular}  &
\begin{tabular}[c]{@{}c@{}}Policy and\\ value level \end{tabular}
  & \begin{tabular}[c]{@{}c@{}}One \\ round \end{tabular}
  & 
  \begin{tabular}[l]{@{}l@{}} SHAPES dataset \\ Traffic junction \\
House3D \\
Predator-prey
\end{tabular}   & - &  \begin{tabular}[c]{@{}c@{}} Spatial memory,\\ scalability  \end{tabular}\\ \hline

I2C \cite{zhu2022survey}  
&  Unicast & \begin{tabular}[c]{@{}c@{}}Existing\\  context \end{tabular}  & \begin{tabular}[c]{@{}c@{}}Weighted\\ linear \\ combination \end{tabular} &
Policy level 
& \begin{tabular}[c]{@{}c@{}}One \\ round \end{tabular}
  & \begin{tabular}[c]{@{}c@{}} Predator-prey  \\Traffic-junction \\ Cooperative navigation \end{tabular}   & \href{https://github.com/PKU-AI-Edge/I2C}{I2C code} & Causal inference \\ \hline

SchedNet \cite{SchedNet}
&   Broadcast  & \begin{tabular}[c]{@{}c@{}}Existing\\  context \end{tabular}  & Concatenation  &
Policy level 
&  \begin{tabular}[c]{@{}c@{}} Multiple  \\ rounds \end{tabular}
  &  \begin{tabular}[c]{@{}c@{}} Predator-prey \\ Starcraft: Broodwars \\Traffic-junction \end{tabular}   & \href{https://github.com/rhoowd/sched_net}{SchedNet code} & Real-world noise \\ \hline

A3C2 \cite{A3C2}
&  Broadcast  & \begin{tabular}[c]{@{}c@{}}Existing\\ context \end{tabular}  & \begin{tabular}[c]{@{}c@{}}Linear \\ combination \end{tabular}  &

\begin{tabular}[c]{@{}c@{}}Policy and\\  value level \end{tabular}
&\begin{tabular}[c]{@{}c@{}} One \\ round \end{tabular}
 &  \begin{tabular}[c]{@{}c@{}}Blind groupUp \\ Navigation \\
Predator-prey\\
Traffic-junction
\end{tabular}   & \href{https://github.com/david-simoes-93/A3C2}{A3C2 Code} & Scalability \\ \hline

MAGIC \cite{zhu2022survey} 
&Unicast  & \begin{tabular}[c]{@{}c@{}}Existing\\ context \end{tabular}  & \begin{tabular}[c]{@{}c@{}}Weighted\\ linear \\ combination \end{tabular}  &

\begin{tabular}[c]{@{}c@{}}Policy and\\  value level \end{tabular}
  &
  \begin{tabular}[c]{@{}c@{}} Multiple  \\ rounds \end{tabular} & 
  \begin{tabular}[c]{@{}c@{}}
Predator-prey\\
Traffic-junction \\
Google research football
\end{tabular}  &  \href{https://github.com/CORE-Robotics-Lab/MAGIC}{MAGIC Code} & Spacial memory \\ \hline

\end{tabular}
\end{table*}
\end{center}

\begin{figure}
    \centering
    \includegraphics[width=1.05\columnwidth]{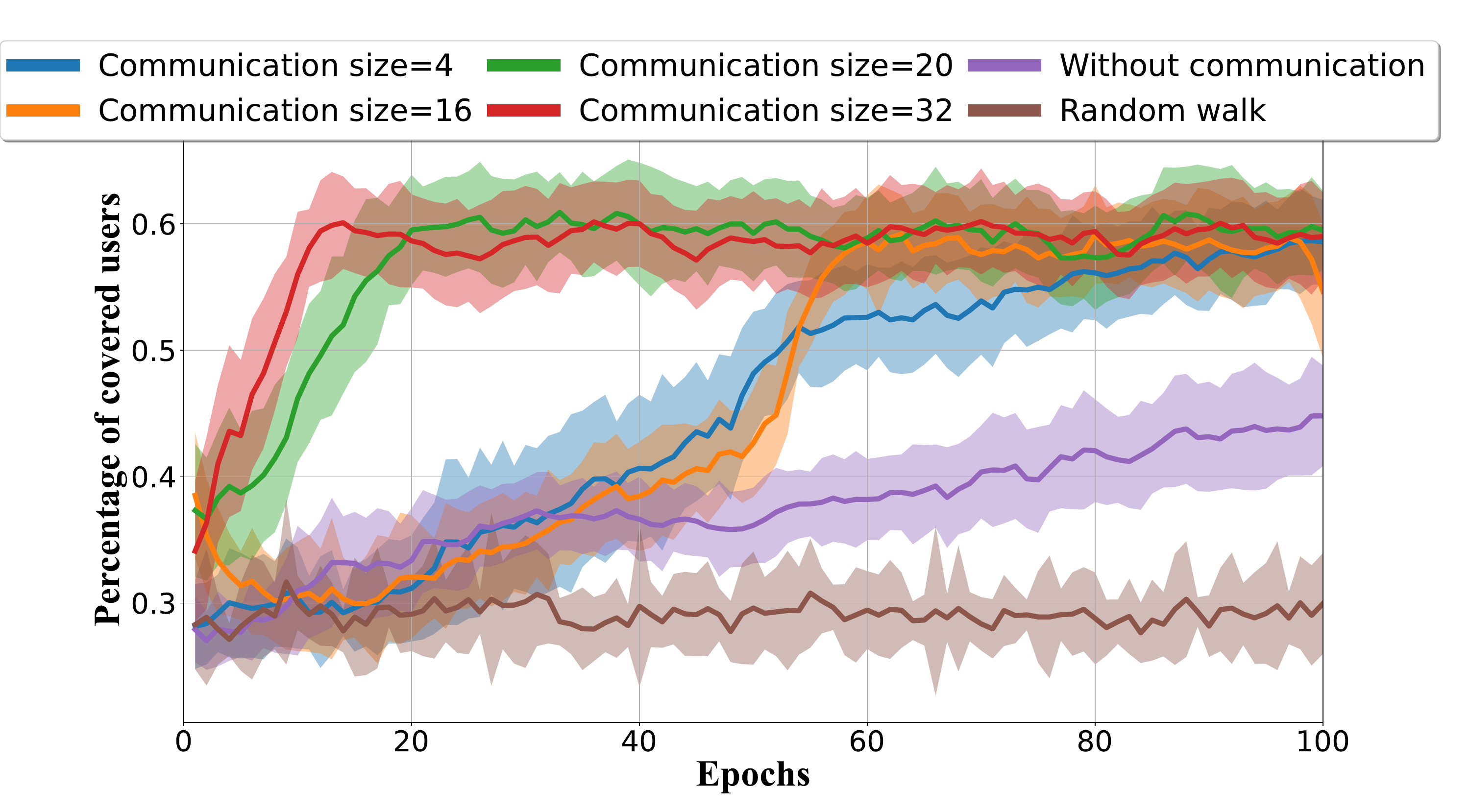}
    \caption{Emergent communication increases the percentage of served users in a flying base station network using the targeted multi-agent communication (TarMAC) algorithm. Three flying base stations are deployed in the environment and try to coordinate in order to find the optimal location for maximizing the coverage for 30 ground users. When learning a communication protocol and exchanging messages, the reward gets higher than when using a classical MARL algorithm without communication or when a random decision-making algorithm such as a random walk is used. Communication size refers to the length of the learned communication message.
    }
    \label{fig:reward}
\end{figure}

\section{Potential Use Cases and Applications} \label{sec:usecases}
There is a significant amount of 6G-enabled emerging use cases and applications that fit into the framework of EC-MARL. The typical scenario is when AI agents are deployed in a partially observable environment to collaboratively solve a complex task. A fully observable environment or a non-rich environment may not require communication or it may be enough to deploy short hard coded communication messages.

\paragraph{\ac{UAV} sector} \acp{UAV} are always limited by a pre-defined flying time which represents a strict energy constraint, that does not allow them to exchange long communication messages. High dimensional messages may be required when the 6G network is deployed at high frequency with a large number of antennas. Therefore, when multiple UAVs are deployed in the field to complete collaboratively a common task, EC-MARL can be a suitable framework to learn jointly the best actions as well as encoded communication messages to be exchanged between agents. A clear application is when UAVs are deployed as flying base stations to efficiently serve users and to maximize the network capacity, each energy-constrained UAV needs to optimize its location depending on its observation (e.g. users' demands, users' location, channel conditions) as well as the messages received by other agents that give information about the non-observable environment. Figure~\ref{fig:reward} shows how the reward increases with the communication size in a simple flying base station setup using parameters of Table~\ref{tab:simulation}. 

\begin{table}[htbp]
\centering
\caption{UAV use case simulation parameters.}
\label{tab:simulation}
\begin{tabular}{@{}lll@{}}
    \toprule
    Parameter & Description & Value \\
    \midrule
    Altitude & UAV flight altitude & 40 \si{\metre} \\
    Speed & UAV flight speed & 45 \si{\metre\per\second} \\
    Maximum Time-step & predefined maximum flight time & 120\\
    Carrier frequency & - & 2 \si{\GHz}\\
    Number of users & - & 120\\
    Number of UAVs & - & 5\\
    \acp{UAV} Capacity & Nb of users served by a \ac{UAV} & 12\\
    Grid size & Served geographical area size & $700 \times 700$ \si{\square\metre} \\
    Transmit power & - & 46 \si{\dB}m\\
    Average noise  power & - & -99 \si{\dB}m\\
    Architecture & EC-MARL architecture & MAGIC \cite{zhu2022survey} \\
    Action pace & Type of actions & Discrete \\
    Number of actions & Space action size & 5 \\
    Discount factor & RL algorithm $\gamma$-discount factor & 0.9 \\
    Number of epochs & - & 100\\
    \bottomrule
\end{tabular}
\end{table}

Collaborative UAVs can also be deployed in field coverage, surveillance and target tracking in military applications for example. In these scenarios, not only energy-efficient communications between UAVs is needed but also it is crucial to communicate the information in a timely manner, which requires learning short communication messages.

\paragraph{Automotive sector} 
the most striking example in the automotive sector is the autonomous driving scenario where vehicles must coordinate with each other to achieve safe driving and fluid traffic. In \cite{A3C2}, an EC-MARL algorithm has been tested in a traffic intersection simulator which is a partially observable environment where multiple intersections are crossed by vehicles. It has been shown through simulations that learned communication between agents makes the traffic fluid without collisions. 
Other applications from the automotive sector would also benefit from the EC-MARL framework such as traffic light control systems, platooning, and fleet management. 
In intelligently connected vehicle networks, vehicles sense their environment and upload these data to the roadside units in order to make decisions about complex tasks such as traffic flow, congestion control, and trajectory planning. In these networks, onboard vehicle sensors capture a large amount of sensing data e.g. images, videos, and traffic situations. Therefore, transmitting raw sensing data is not viable when a large number of vehicles are deployed, and emergent communication in this MARL framework can make such networks more efficient.

\paragraph{Smart factories and robotics} in these environments where complex tasks need to be completed such as automotive production, navigation, and control tasks, EC-MARL allows low latency communications by encoding efficiently exchanged communication messages.
In multi-agent navigation, agents that need to complete cooperative navigation tasks have been able to learn emergent communication protocols when deployed in different grid-world environments, and have successfully completed the navigation task
\cite{A3C2}.
In addition, in disaster scenes such as buildings on fire, earthquakes, wars, agents need to coordinate between them in order to cover and search the whole area for rescue missions. Such environments are hostile, constantly changing, and agents need to communicate between them to complete this complex task. Sometimes the target may be moving, which makes the task even more complex. An example of such search and rescue missions has been studied in \cite{Tarmac} using the House3D framework. 

In addition, in \cite{A3C2}, EC-MARL has been tested on a noisy learned communication and has been shown to be robust against noise. Emergent communication may be more robust against noise and interference, which enhances reliability in 6G networks.

\section{Challenges and Opportunities} \label{sec:challenges}
Emergent communication in MARL systems and its integration into 6G wireless networks is a nascent and complex field that raises several open research questions and opportunities. Next, we present some of the main challenges that will make an impact on 6G-driven applications.

\paragraph{Scalability} a large amount of work on emergent communications in the literature focuses on small-scale problems, which may not reflect 6G network models that involve a large number of devices and a large amount of sensing data, and may hinder the generality of the presented conclusions. In \cite{chaabouni2021emergent}, a preliminary study has been conducted to show the effect of scaling up the population size, the dataset size as well as the task complexity. Several challenges emerge such as the instability of the learning process and the generality and guarantees of the performance. Mean field theory is one direction to explore to handle a large number of agents.

\paragraph{Non-stationarity} In realistic communication scenarios, environment dynamics and rewards are not always stationary, and since we assume stationarity in our learning algorithms, they may not perform well in practical applications. Developing EC-MARL algorithms that account for the non-stationarity of the environment is an interesting challenge to investigate e.g. lifelong learning systems. 
\paragraph{Convergence} even for relatively simple environments, algorithms of MARL with communications need an important number of iterations before converging. Investigating methods to speed up the convergence would be beneficial to save computational resources, especially when deployed in energy-constrained agents such as \acp{UAV}. One way that may help speed up the convergence during the learning phase is the share of prior knowledge of the environment and the agents. Self-supervised learning, federated learning, and transfer learning may also improve model convergence. Straggler's problem can also significantly delay the learning process and the convergence and needs to be addressed.

\paragraph{Measuring the effectiveness of emergent communication} the challenge is to know whether exchanged messages affect the actions taken by the agents, and to quantify this impact. To study the effectiveness of emergent communication in complex environments, it is not sufficient to show that adding a communication channel leads to an increased reward \cite{lowe2019pitfalls}. The open challenge is to define metrics in order to evaluate the impact of an agent's message on another agent's action. Some metrics have been defined in \cite{lowe2019pitfalls}, but no metric captures the full insight behind the agents' behaviors.
\paragraph{Effect of noisy communications} most of the emergent communication work do not consider realistic communication settings with a practical channel model and interference. In \cite{A3C2}, sources of noise have been considered, but a more realistic communication chain needs to be considered and the effect on the emergent language needs to be studied.
\paragraph{Communication cost} Communication is costly, and exchanging messages among agents can be expensive in some scenarios. The value of communication should make up for its cost. The trade-off between communication cost and partial observation should be studied to identify in which cases and how much communication is needed.

\paragraph{Understanding semantics of emergent communication} agents develop a communication protocol they understand, but as an external observer that has not been involved in the interaction, it is difficult to decode the messages and extract their meaning. We may only guess what the messages might be referring to. This also opens another challenge of evaluating machine's communication with humans. \cite{andreas2017translating} has proposed to interpret the induced communication messages by translating them to human language in order to facilitate collaboration between humans and machines,  another possible way is to encourage agents to learn messages similar to natural languages \cite{lazaridou2016multi}. 
\paragraph{Compositionality,  generalization and interoperability of the communication protocol} Can agents refer to new composite meaning e.g. \emph{high temperature}, once they learned separate concepts e.g. \emph{high} and \emph{temperature}? Can the learned protocol generalize to understand new concepts based on learning specific concepts? How can we teach a new agent a communication protocol to be able to communicate with other agents?
\paragraph{Learning a well-structured communication protocol} in emergent communication, agents learn messages that complete the task without imposing efficient encoding. In \cite{chaabouni2019anti}, it has been shown that, unlike human language where more frequent words are efficiently associated with shorter words, agents can develop an anti-efficient encoding in emergent communication, where the most frequent inputs are associated with the longest messages, which make them easier to discriminate.
\paragraph{Emergent communication with contextual reasoning} if agents can send messages to themselves, which will create a recurrent network, then they can improve messages sent to other agents through contextual reasoning. \cite{seo2021semantics} shows that by iteratively reasoning about the communication context of the listener agent, the length of the semantic representation of the message to transmit can be reduced significantly.

\section{Conclusion} \label{conclusions}

This article underscores the importance and relevance of EC-MARL for designing future 6G wireless networks, and shows the potential of learning communication protocols. We have shown that multiple 6G-driven applications can fit into this framework. Since these algorithms have been tested by the ML community in simple and controlled simulated environments, we have highlighted several research opportunities specific to the integration of EC-MARL into 6G systems, which opens new research directions to the communication research community.

\bibliographystyle{IEEEtran}
\bibliography{ref}

\begin{thebibliography}{10}
\providecommand{\url}[1]{#1}
\csname url@samestyle\endcsname
\providecommand{\newblock}{\relax}
\providecommand{\bibinfo}[2]{#2}
\providecommand{\BIBentrySTDinterwordspacing}{\spaceskip=0pt\relax}
\providecommand{\BIBentryALTinterwordstretchfactor}{4}
\providecommand{\BIBentryALTinterwordspacing}{\spaceskip=\fontdimen2\font plus
\BIBentryALTinterwordstretchfactor\fontdimen3\font minus
  \fontdimen4\font\relax}
\providecommand{\BIBforeignlanguage}[2]{{%
\expandafter\ifx\csname l@#1\endcsname\relax
\typeout{** WARNING: IEEEtran.bst: No hyphenation pattern has been}%
\typeout{** loaded for the language `#1'. Using the pattern for}%
\typeout{** the default language instead.}%
\else
\language=\csname l@#1\endcsname
\fi
#2}}
\providecommand{\BIBdecl}{\relax}
\BIBdecl

\bibitem{10041914}
M.~Chafii, L.~Bariah, S.~Muhaidat, and M.~Debbah, ``{Twelve Scientific
  Challenges for 6G: Rethinking the Foundations of Communications Theory},''
  \emph{IEEE Communications Surveys \& Tutorials}, vol.~25, no.~2, pp.
  868--904, 2023.

\bibitem{lazaridou2020emergent}
A.~Lazaridou and M.~Baroni, ``Emergent multi-agent communication in the deep
  learning era,'' \emph{arXiv preprint arXiv:2006.02419}, 2020.

\bibitem{seo2021semantics}
H.~Seo, J.~Park, M.~Bennis, and M.~Debbah, ``{Semantics-native communication
  with contextual reasoning},'' \emph{arXiv preprint arXiv:2108.05681}, 2021.

\bibitem{zhu2022survey}
C.~Zhu, M.~Dastani, and S.~Wang, ``{A Survey of Multi-Agent Reinforcement
  Learning with Communication},'' \emph{arXiv preprint arXiv:2203.08975}, 2022.

\bibitem{foerster2016learning}
J.~Foerster, I.~A. Assael, N.~De~Freitas, and S.~Whiteson, ``{Learning to
  communicate with deep multi-agent reinforcement learning},'' \emph{Advances
  in neural information processing systems}, vol.~29, 2016.

\bibitem{IntensionSharing}
W.~Kim, J.~Park, and Y.~Sung, ``{Communication in multi-agent reinforcement
  learning: Intention sharing},'' in \emph{International Conference on Learning
  Representations}, 2020.

\bibitem{actorAttentionCritic}
S.~Iqbal and F.~Sha, ``{Actor-attention-critic for multi-agent reinforcement
  learning},'' in \emph{International Conference on Machine Learning}.\hskip
  1em plus 0.5em minus 0.4em\relax PMLR, 2019, pp. 2961--2970.

\bibitem{Tarmac}
A.~Das, T.~Gervet, J.~Romoff, D.~Batra, D.~Parikh, M.~Rabbat, and J.~Pineau,
  ``{Tarmac: Targeted multi-agent communication},'' in \emph{International
  Conference on Machine Learning}.\hskip 1em plus 0.5em minus 0.4em\relax PMLR,
  2019, pp. 1538--1546.

\bibitem{SchedNet}
D.~Kim, S.~Moon, D.~Hostallero, W.~J. Kang, T.~Lee, K.~Son, and Y.~Yi,
  ``{Learning to schedule communication in multi-agent reinforcement
  learning},'' \emph{arXiv preprint arXiv:1902.01554}, 2019.

\bibitem{A3C2}
D.~Sim{\~o}es, N.~Lau, and L.~P. Reis, ``{Multi-agent deep reinforcement
  learning with emergent communication},'' in \emph{2019 International Joint
  Conference on Neural Networks (IJCNN)}.\hskip 1em plus 0.5em minus
  0.4em\relax IEEE, 2019, pp. 1--8.

\bibitem{chaabouni2021emergent}
R.~Chaabouni, F.~Strub, F.~Altch{\'e}, E.~Tarassov, C.~Tallec, E.~Davoodi,
  K.~W. Mathewson, O.~Tieleman, A.~Lazaridou, and B.~Piot, ``{Emergent
  communication at scale},'' in \emph{International Conference on Learning
  Representations}, 2021.

\bibitem{lowe2019pitfalls}
R.~Lowe, J.~Foerster, Y.-L. Boureau, J.~Pineau, and Y.~Dauphin, ``{On the
  pitfalls of measuring emergent communication},'' \emph{arXiv preprint
  arXiv:1903.05168}, 2019.

\bibitem{andreas2017translating}
J.~Andreas, A.~Dragan, and D.~Klein, ``{Translating neuralese},'' \emph{arXiv
  preprint arXiv:1704.06960}, 2017.

\bibitem{lazaridou2016multi}
A.~Lazaridou, A.~Peysakhovich, and M.~Baroni, ``{Multi-agent cooperation and
  the emergence of (natural) language},'' \emph{arXiv preprint
  arXiv:1612.07182}, 2016.

\bibitem{chaabouni2019anti}
R.~Chaabouni, E.~Kharitonov, E.~Dupoux, and M.~Baroni, ``{Anti-efficient
  encoding in emergent communication},'' \emph{Advances in Neural Information
  Processing Systems}, vol.~32, 2019.

\end{thebibliography}

\end{document}